\documentclass[10pt,twocolumn,letterpaper]{article}

\usepackage{iccv}
\usepackage{times}
\usepackage{graphicx}
\usepackage{amsmath}
\usepackage{amssymb}
\usepackage{url}
\usepackage{pdfpages}

\usepackage[breaklinks=true,bookmarks=false]{hyperref}

\usepackage[utf8]{inputenc}
\usepackage{authblk}

\iccvfinalcopy 

\def\iccvPaperID{2830}

\ificcvfinal\fi

\begin{document}

\title{Tag2Pix: Line Art Colorization Using Text Tag With SECat and Changing Loss}

\makeatletter
\newcommand{\printfnsymbol}[1]{
  \textsuperscript{\@fnsymbol{#1}}
}
\makeatother

\author{Hyunsu Kim\thanks{equal contribution}}
\author{Ho Young Jhoo\printfnsymbol{1}}
\author{Eunhyeok Park}
\author{Sungjoo Yoo}

\affil{Seoul National University
 \authorcr
  \{\tt gustnxodjs, mersshs, eunhyeok.park, sungjoo.yoo\}@gmail.com}

\maketitle
\ificcvfinal\thispagestyle{empty}\fi

\begin{abstract}

Line art colorization is expensive and challenging to automate. A GAN approach is proposed, called Tag2Pix, of line art colorization which takes as input a grayscale line art and color tag information and produces a quality colored image. First, we present the Tag2Pix line art colorization dataset. A generator network is proposed which consists of convolutional layers to transform the input line art, a pre-trained semantic extraction network, and an encoder for input color information. The discriminator is based on an auxiliary classifier GAN to classify the tag information as well as genuineness. In addition, we propose a novel network structure called SECat, which makes the generator properly colorize even small features such as eyes, and also suggest a novel two-step training method where the generator and discriminator first learn the notion of object and shape and then, based on the learned notion, learn colorization, such as where and how to place which color. We present both quantitative and qualitative evaluations which prove the effectiveness of the proposed method.

\end{abstract}

\section{Introduction}
Line art colorization is an expensive, time-consuming, and labor-intensive task in the illustration industry. It is also a very challenging task for learning methods because the output is a fully colorized image, but the only input is a monotone line art and a small amount of additional information for colorization (\eg, color strokes). The multimodal learning of segmentation and colorization is essential.

Recently, various studies have been conducted on colorization. Most of those works are based on generative adversarial network (GAN)~\cite{gan}, and we focus on colorization using text and line art. In researches on text hint colorization, some works tried to colorize a grayscale image with information given by a text sentence describing the color of each object~\cite{chen2018language,1804.06026}, while others modified the color of a particular part of the image using sentences~\cite{sisgan,nam2018text}. Though several studies exist for text-based colorization, none of them have focused on line art colorization, which is more difficult due to the relatively low amount of information contained in the input image.

In the case of line art colorization, there are two typical ways to give hints for colorization. In user-guided colorization~\cite{style2paints,scribbler,paintschainer,zhang2018two}, short lines with the desired color are drawn over target locations on the line art, and the output is generated by naturally filling the remaining space. In the style-transfer method~\cite{comicolor, mangacolorsingle, style2paints,StyleTransfer}, an existing sample image is used as a hint for the generative network, and the output is generated following the color distribution of the given sample image. These methods successfully simplify the colorization process, but still require intervention through either skilled professionals (user-guided case) or images with similar patterns (style-transfer case), both of which are expensive.

\begin{figure}[ht]
   \centering
   \includegraphics[width=0.8\columnwidth]{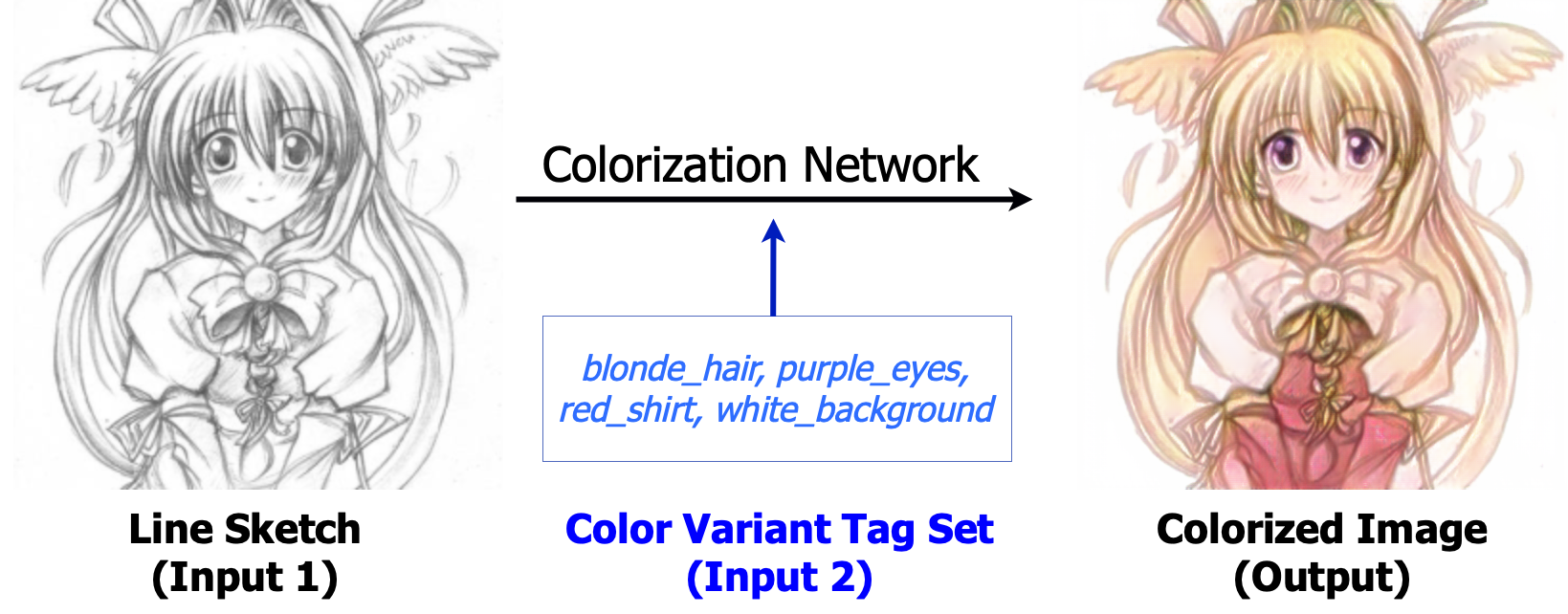}
   \caption{Example of tag-based colorization method.}
   \label{img:figure1}
\end{figure}

As an alternative, we define a new problem for implementing line art colorization based on tag data, as shown in Figure~\ref{img:figure1}. The generator network receives input data as a monotone line art and a corresponding color variant tag (CVT), such as \textit{blue\_hair} or \textit{red\_eyes}, after which the network uses these two input data to colorize the monotone line art based on the given tag information. This tag-based method minimizes the effort of offering hints for colorization; thus it can provide high quality colorization output without requiring the intervention of a skilled professional.

In this paper, we propose a GAN approach to tag-based line art colorization as well as a novel architecture for multi-label segmentation and colorization. We also present a novel training scheme which improves the stability of GAN training as well as its output quality. In addition, we offer our source codes with accompanying pre-trained networks so that readers can reproduce the results.\footnote{https://github.com/blandocs/Tag2Pix} Our contributions are as follows:

\begin{itemize}
      \item \textbf{Tag2Pix dataset} \\
      We offer a dataset for training the Tag2Pix network. The dataset is composed of four sets of data: color illustrations, monotone line arts, color invariant tags (CITs), and CVTs.
      
      \item \textbf{Tag2Pix network} \\
      We present an adversarial line art colorization network called Tag2Pix. This network is a variation of auxiliary classifier GAN (ACGAN)~\cite{acgan} and designed to generate a colorized image based on a monotone line art and CVTs.
      
      \item \textbf{SECat: Squeeze and Excitation with Concatenation} \\
      We propose a novel network structure that enhances multi-label segmentation and colorization. This method helps to color even small areas such as eyes. 

      \item \textbf{Two-step training with changing loss} \\
      We present a novel loss combination and curriculum learning method for the Tag2Pix network. This method divides the learning focus between segmentation and colorization in order to train the network in a stable and fast manner.
\end{itemize}

\section{Related work}

\subsection{GAN for colorization}

GAN~\cite{gan} offers superior quality in generation tasks compared to conventional image generation methods. Pix2pix~\cite{pix2pix} changes grayscale images to color images by adding GAN’s adversarial loss to the reconstruction loss typically used for CNN. ACGAN~\cite{acgan} and Zhang \etal~\cite{StyleTransfer} show that an auxiliary decoder network can increase the stability and expression power of a generative network. In this paper, this aspect of ACGAN is utilized in the design of our network because it is suitable for learning the features of CVTs and CITs.

\subsection{Sketch-based line art colorization}

Several studies have been conducted on GAN for line art colorization. A typical method of line art colorization is to provide a color hint for the line art. PaintsChainer~\cite{paintschainer}, Ci \etal~\cite{colorstroke}, and Scribbler~\cite{scribbler} implement automatic colorization by using short lines of specified color as hints for the colorization of targeted line art areas. In StyleTransfer~\cite{StyleTransfer}, the color style of an illustration used as input data is transferred to the original line art image. Style2Paints~\cite{style2paints,zhang2018two} extends StyleTransfer by adding a refinement stage, which provides a state-of-the-art result. However, both methods still require expertise from the user for either specifying colors at every line-separated location, or for preparing a properly styled illustration as input data for each original line art. Due to these limitations, colorization remains expensive and difficult. In this work, we propose a simple tag-based colorization scheme to provide a low-cost and user-friendly line art colorization option.

\subsection{Text related vision task}

StackGAN~\cite{stackgan} accepts full text sentences to synthesize a color image. SISGAN~\cite{sisgan} and Nam \etal~\cite{nam2018text} use sentences to change the color of a particular area
in a color image. Manjunatha \etal~\cite{1804.06026} and Chen \etal~\cite{chen2018language} colorize grayscale images using sentences that describe objects with colors. Meanwhile, Illustration2vec~\cite{i2v} uses a VGG~\cite{vgg} network to extract semantic tags from color illustration. Based on Illustration2vec, Jin \etal~\cite{mgmsite,makegirlsmoe} present a study on creating artificial anime-style faces by combining a given tag with random noise. The concept of extracting semantic features using CNN is adopted in our work to extract CITs from a given line art.

\subsection{Rebalance weights in feature maps}

In styleGAN~\cite{stylegan}, a style-based generator is used to improve the output quality by modifying intermediate features after convolutional layers based on adaptive instance normalization (AdaIN~\cite{adain}). By injecting the encoded style information into the intermediate feature maps, the generator can synthesize realistic images for various target styles. In this work, we propose a novel network structure called SECat, which improves the stability and quality of line art colorization with minimal effort.

\section{Problem and Dataset}
\subsection{Problem definition}

Our problem is a tag-based line art colorization which automatically colors a given line art using color variant text tags provided by the user. Line art is a grayscale image that includes only the edges of objects, and a CVT determines the desired color of the target object in the line art. In order to address this problem, our proposed network extracts a feature, such as \textit{hat} or \textit{backpack}, which provides information about the shape of the image for color invariant tags (CITs) in a given line art. Thus, as shown in Figure~\ref{img:figure1}, our proposed network colorizes a given line art with color variant tags and color invariant tag features. Note that the user does not need to provide color invariant tags, and only has to provide the color-related tags.

\subsection{Tag2Pix dataset}
\label{section_3_2}
\paragraph{Data filtering}
We used a large-scale anime style image dataset, Danbooru2017~\cite{danbooru2017}. Each illustration of Danbooru2017 has very detailed tags for pose, clothing, hair color, eye color, and so on. In order to achieve our goals, we selected $370$ CITs and $115$ CVTs that were each used more than $200$ times for our training dataset.

We also selected images with only one person in simple background.
Each image used from the Danbooru2017 set has a size of ${512\times512}$ letterboxed. Through mirror-padding, we removed the letterbox and obtained $39,031$ high-definition images. Additional face images were also extracted through lbpascade\_animeface~\cite{animeface} to provide more delicate face colorization. A total of $16,224$ face images were obtained by selecting images with a resolution of ${128\times128}$ or higher.

\paragraph{Line art extraction}

\begin{figure}[ht]
    \centering
  \includegraphics[width=\columnwidth]{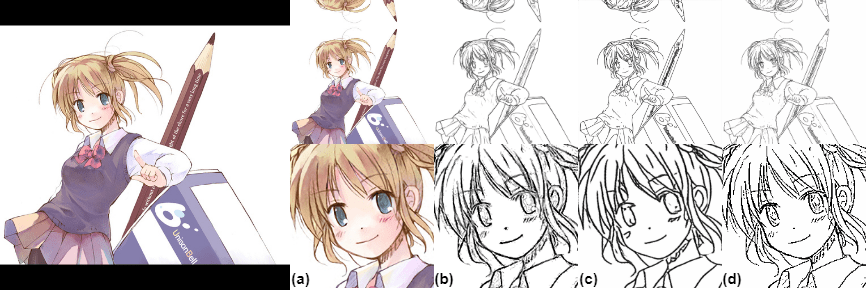}
  \caption{Example of line art extraction methods. Left: sample image from Danbooru2017. Top: training data obtained by (a) removing letterbox with mirror padding, (b) sketchKeras~\cite{sketchkeras}, (c) Sketch Simplification~\cite{simplification}, and (d) XDoG~\cite{xdog}. Bottom: cropped face images using lbpcascade\_animeface~\cite{animeface}.} 
  \label{img:line_extraction}
\end{figure}

We needed to extract line arts from color illustrations for supervised learning, but traditional edge detection algorithms, such as Canny Edge Detector~\cite{canny}, failed to create natural artistic line arts. Thus, in order to obtain line arts with clear edges and various styles, we used multiple methods of line art extraction as illustrated in Figure~\ref{img:line_extraction}.

We mainly extracted sketches through the sketchKeras~\cite{sketchkeras} network specialized in line art creation in the style of anime. However, the sketchKeras images showed a tendency toward the style of a pencil sketch, with the presence of an unnecessary depiction or incorrect control of the line thickness. We therefore used a simplified sketch based on sketchKeras, additionally using Sketch Simplification~\cite{simplification, simple0}. These images are close to digital line art because they have nearly constant line thickness. Because the line thickness of the simplified result depends on the resolution of the input image, the input was enlarged to ${768\times768}$. Lastly, we extracted an algorithmic line art from the grayscale of the original color image using XDoG~\cite{xdog}.

Training a network with only a single type of sketch, such as sketchKeras, has shown a tendency to overfit to the sketch input and retrace the RGB values from the sketch.  Through various styles of sketches, we have been able to avoid this effect.

\section{Tag2Pix network and loss design}
\subsection{Generator and discriminator network}

\begin{figure*}[ht]
  \centering
  \includegraphics[width=1.0\textwidth]{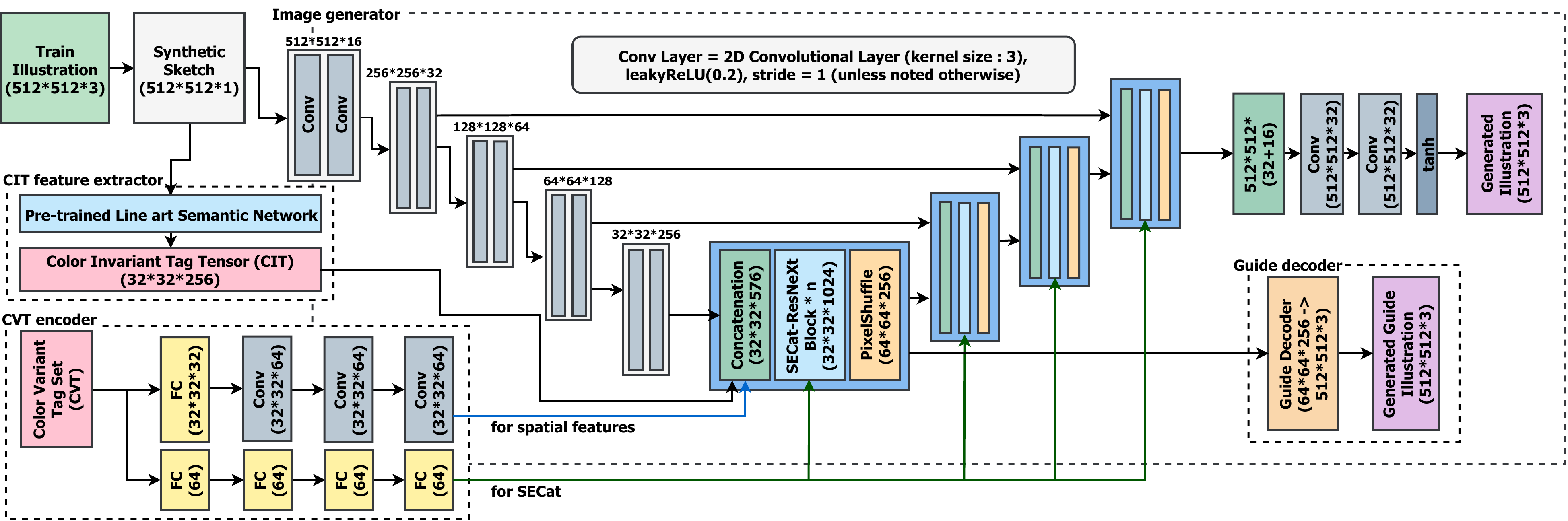}
  \caption{ The overall structure of the generator network. The numbers within brackets represent tensor dimension (width$\times$height$\times$depth). }
  \label{img:generator}
\end{figure*}

Figure~\ref{img:generator} shows the architecture of the generator network, which takes a line art and color tags as input and gives a colored illustration as output. As shown in Figure~\ref{img:generator}, the network is composed of four main parts: CIT feature extractor, image generator, CVT encoder, and guide decoder. The user can obtain a colored line art simply by providing a line art and CVTs without providing CITs, as shown in Figure~\ref{img:generator}.

CITs tend to represent shape information which can serve as a guide for better coloring, providing useful information for the correct applicaiton of color to desired positions. The CIT feature extractor is designed to extract the features of CITs from the given line art, and to utilize those in our proposed GAN architecture. In order to extract the features of CITs, we pre-trained a feature extraction network based on SE-ResNeXt-50~\cite{senet,resnext}. The network was trained to predict the multi-label CITs and the intermediate feature map after \textit{conv3\_4} with ReLU is provided to the image generator.

The CVT encoder is designed to embed the given CVTs into the latent space. The CVT encoder consists of two sets of layers for output, providing spatial features to be merged into the encoded feature map of image generator, and assisting the layer re-calibration feature of SECat. SECat is explained in Section~\ref{section_4_2}. The CVT input is first encoded as a one-hot vector, and the vector is separately embedded through multiple fully-connected (FC) layers and convolution layers. Even though CVTs do not have spatial information, convolution layers have better performance than FC layers in lower computation overhead.

The image generator is based on U-Net~\cite{unet}, which is designed to generate high-resolution images. As Figure~\ref{img:generator} shows, the image generator first produces an intermediate representation ($32 \times 32 \times 576$) by concatenating the feature maps of convolutional layers ($32 \times 32 \times 256$) with both the CIT feature extractor output and the spatial features output of the CVT encoder for U-Net. The multiple decoder blocks then run to produce a high-quality colored illustration. Specifically, each decoder block takes as input the feature maps from the previous decoder block and the convolutional layer of the same spatial dimension in the U-Net structure. Each decoder block is based on the pixel shuffle operation, which is an upsampling method used to reduce checkerboard artifacts~\cite{pixelshuffle}.

The image generator has a deep structure, and can result in the vanishing gradient problem. In order to facilitate network training, we adopt the guide decoder~\cite{StyleTransfer}. As shown in Figure~\ref{img:generator}, the guide decoder is connected to the first decoder block, and produces a colored illustration. The guide decoder offers a new loss path to an intermediate feature map, which improves quality and helps to mitigate the vanishing gradient problem.

\begin{figure}[ht]
  \includegraphics[width=\columnwidth]{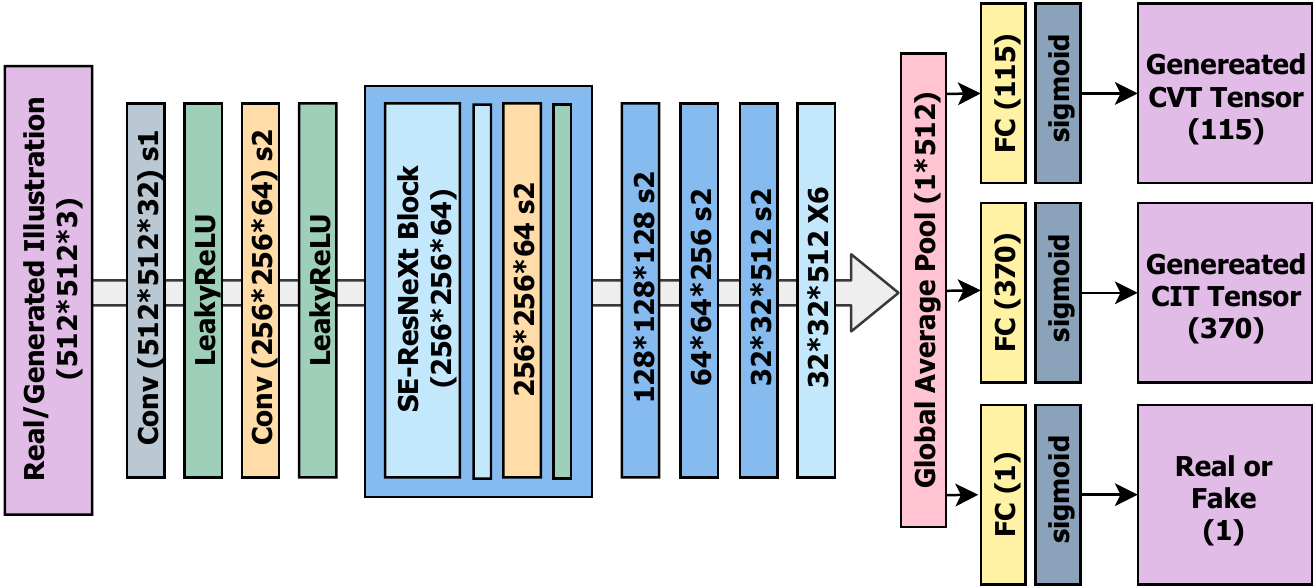}
  \caption{The overall structure of the discriminator network.}
  \label{img:discriminator}
\end{figure}

Figure~\ref{img:discriminator} shows the overview of the discriminator. After receiving a color image as input, the discriminator determines whether the input is genuine, and at the same time predicts which CVTs and CITs are used.

This network is inspired by ACGAN~\cite{acgan}, as the multi-label classification (especially for CVTs) plays a key role in the target task. The generator and discriminator of the existing ACGAN are trained to successfully generate images of a single class. On the contrary, our GAN is trained to generate colored illustrations and multi-label classification using CVTs and CITs.

\subsection{SECat - Squeeze and Excitation with Concatenation}
\label{section_4_2}

Generally, colorization hints are encoded into a high-level feature which is provided to the generator network as an input to the decoder block~\cite{colwithword,colorstroke,StyleTransfer}. However, according to our observation, the aforementioned method has a disadvantage for tag-based line art colorization. When we adopt this method for the proposed network by encoding the CVT input into a spatial feature and merging it with the input feature of the generator’s decoder block, colorization is performed well for large objects such as hair, but not for the detailed small objects such as eyes.

In order to overcome this limitation, we propose a novel structure named SECat (Squeeze and Excitation with Concatenation). SECat is inspired by the network structure of styleGAN~\cite{stylegan}, which adjusts the intermediate feature map using affine transform in order to inject style information to the generator with similar structure to SENet~\cite{senet} in terms of weight rebalancing.

\begin{figure}[ht]
  \includegraphics[width=\columnwidth]{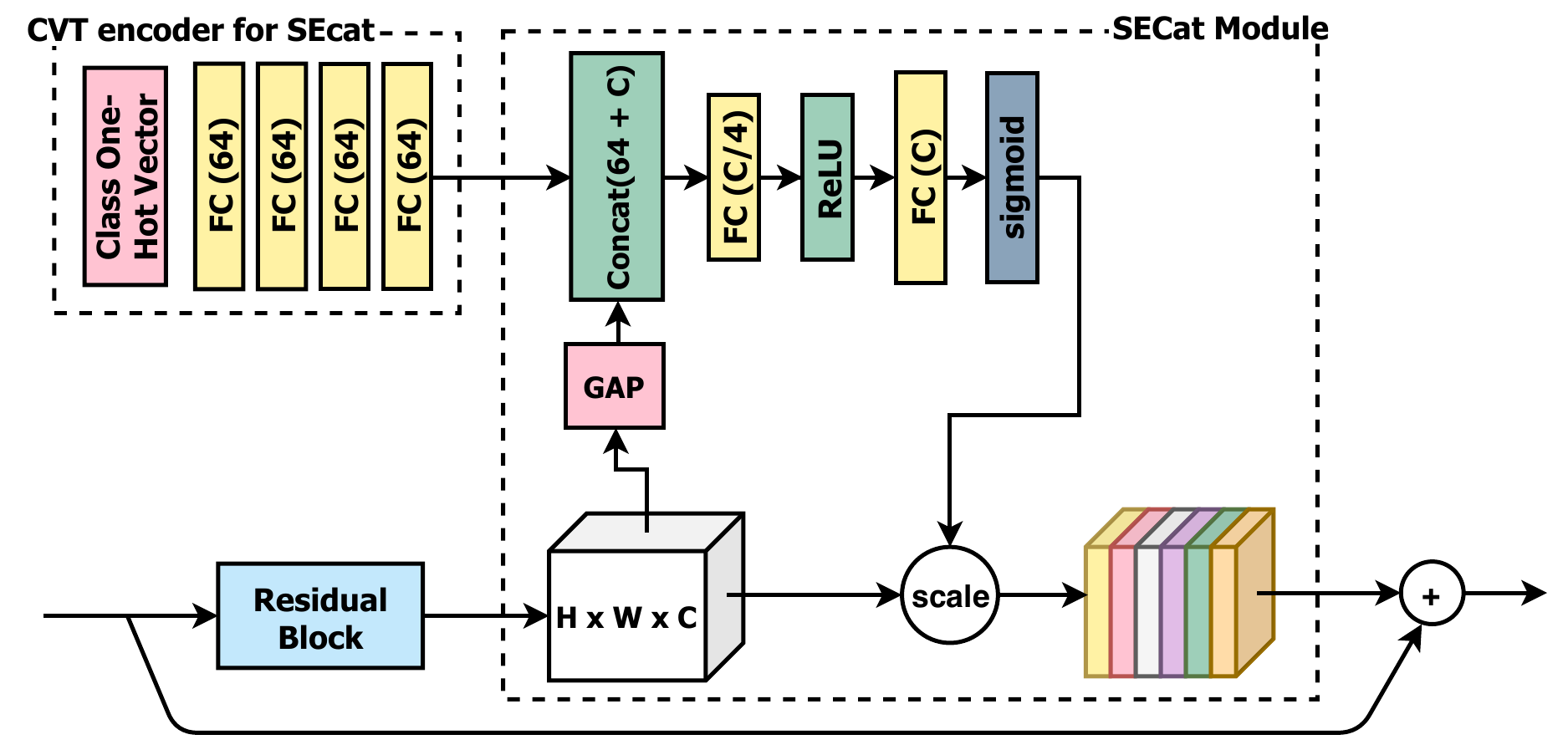}
  \caption{SECat block.}
  \label{img:SECat}
\end{figure}

Figure~\ref{img:SECat} shows an overall structure of SECat. In SENet~\cite{senet}, the \textit{squeeze} is the global information extracting process that generates a channel-dimension vector using 2-D spatial global average pooling. \textit{Excitation} is the adaptive recalibration process that generates channel-wise importance weights using two FC layers with a bottleneck. The output feature maps of a residual block are scaled by channel-wise importance weights obtained through a squeeze and excitation process. In the case of SECat, the user-given CVTs are encoded by several FC layers, and the encoded vectors are merged to the output feature of the squeeze process. This merged feature is propagated to the excitation process so that CVT information is not only incorporated, but also utilized to emphasize important features in all decoding blocks of the generator.

\subsection{Loss design with two-step curriculum training}

\begin{figure}[ht]
  \centering
  \includegraphics[width=0.75\columnwidth]{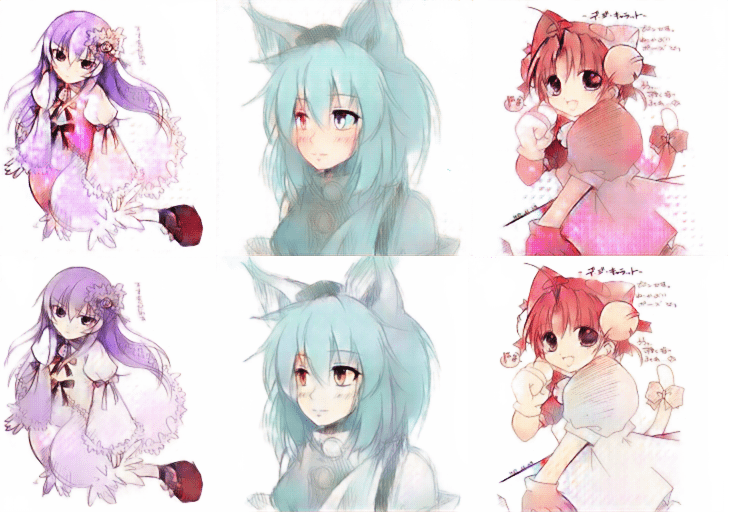}
  \caption{Illustration of two-step training effect. Top: single-step trained result, Bottom: two-step trained result. In the single-step case, color is mixed and bled in each segment.}
  \label{img:2step_effect}
\end{figure}

Unlike previous works which utilize hints containing both color and location information~\cite{colorstroke,paintschainer}, a tag-based hint does not contain any information of location and shape to guide coloring. Thus, semantic segmentation for localized colorization is especially difficult to perform due to the lack of spatial information and RGB value hints. However, if the network is trained for both tasks in a single stage, the generated images often suffer from the problems of color mixing and bleeding, as exemplified in Figure~\ref{img:2step_effect}. According to our observation, when training the network for both segmentation and coloring, it is essential for the network to learn two tasks in the curriculum, first learning semantic segmentation, followed by colorization.

In order to obtain the curriculum-based learning, we propose a two-step training with changing loss where learning proceeds while applying two different losses sequentially.

\paragraph{Step 1. Segmentation}

In the first step, we focus on low-level features like edge and shade. The tag classification losses are not utilized for both generator and discriminator; thus the generator and discriminator are trained based only on adversarial loss $\mathcal{L}_{adv}$ and reconstruction loss $\mathcal{L}_{rec}$ as follows:

{
\small
\begin{equation} \label{step1_loss}
\begin{split}
    &\mathcal{L}_D = -\mathcal{L}_{adv}^{}, \\
    &\mathcal{L}_G = \mathcal{L}_{adv}^{} + \lambda_{rec} \mathcal{L}_{rec}^{}.
\end{split}
\end{equation}
}

The more detailed equation of adversarial loss is as follows:

{
\small
\begin{equation}
\begin{split}
\mathcal{L}_{adv}^{}=
\mathbb{E}_{y}[\log D_{adv}(y)] + \mathbb{E}_{x}[\log(1 - D_{adv}(G_f(x,c_v)))],
\end{split}
\end{equation}
}

where $x$ and $y$ are the paired domain of line arts and real color illustrations, $c_v$ is the CVT values, $G_f$ is the synthesized color image from generator, and $D_{adv}$ is the discriminator output of real or fake. In Equation~\ref{step1_loss}, $\lambda_{rec}$ is a weighting factor for the reconstruction loss $\mathcal{L}_{rec}$ that is expressed as follows:

{
\small
\begin{equation} \label{l1_loss}
    \mathcal{L}_{rec} = \mathbb{E}_{x,y}[|| y - G_f (x, c_v)||_{1} + \beta || y - G_g (x, c_v)||_{1}],  
\end{equation}
}

where $G_g$ is the output of the guide decoder and hyperparameter $\beta$ is $0.9$. $\mathcal{L}_{rec}$ is a pixel-wise L1 loss between an authentic and the generated images. We set $\lambda_{rec}$ large enough ($1000$) to make the network to follow the original image distribution rather than deceiving each other. This training stage guides the network to learn semantic information regarding line arts, resulting in a more precise and clearer borderline with adequate light and shade.

\paragraph{Step 2. Colorization}

In the second step, the tag classification loss $\mathcal{L}_{cls}$ is introduced, which makes the generator and discriminator learn colorization based on a better understanding of object shape and location. The step 2 losses are as follows:

{
\small
\begin{equation} \label{step2_loss}
\begin{split}
    &\mathcal{L}_D = -\mathcal{L}_{adv} + \lambda_{cls} \mathcal{L}_{cls}, \\
    &\mathcal{L}_G = \mathcal{L}_{adv} + \lambda_{cls} \mathcal{L}_{cls} + \lambda_{rec} \mathcal{L}_{rec}.      
\end{split}
\end{equation}
}

In Equation~\ref{step2_loss}, $\lambda_{cls}$ is the weighting factor indicating the importance of the CIT and CVT. Note that the weighting factor plays an important role in our proposed two-step training with changing loss. The classification loss, which trains the discriminator for evaluating the amount of information associated with tags in the input image, is obtained as follows:

{
\small
\begin{equation} \label{g_class}
\begin{split}
    \mathcal{L}_{cls} = \mathbb{E}_{y,c_v,c_i}[-\log D_{cls}(c_v,c_i|y)] + \\
    \mathbb{E}_{x,c_v,c_i}[-\log D_{cls}(c_v,c_i|G_f (x, c_v))],
\end{split}
\end{equation}
}

where $c_i$ is the CIT values and $D_{cls}(c_v,c_i|y)$ binary classification for each tag, given $y$. The discriminator tries to predict CVT and CIT with a high probability. According to our experiments, the two-step approach makes the training of our GAN model more stable and faster.

\paragraph{Colorization for real-world sketch}

\begin{figure}[ht]
  \centering
  \includegraphics[width=\columnwidth]{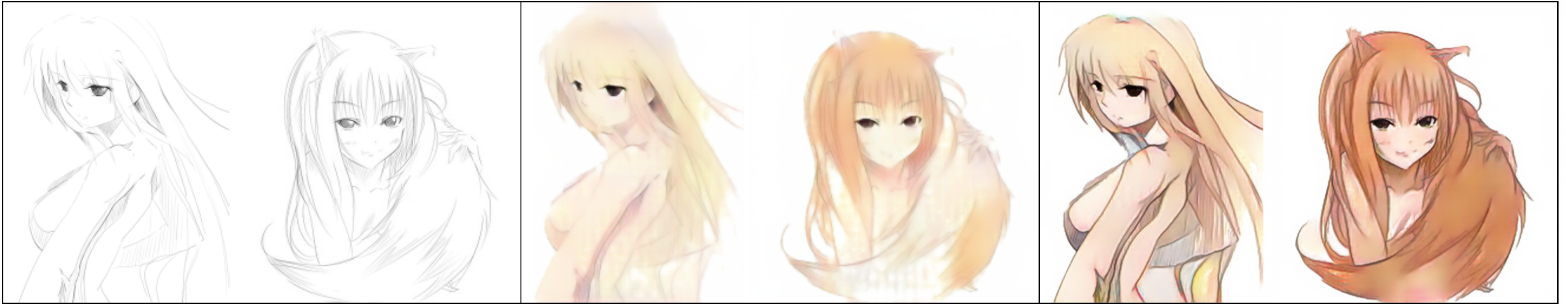}
  \caption{Real-world line arts (left), colorization results without (middle) and 
with brightness adjustment (right).}
  \label{img:brightness}
\end{figure}

Although well-colored results can be obtained with two-step training, but a problem remains, as our network tends to be overfitted to our artificial sketch input (Section~\ref{section_3_2}). The artificial sketch becomes very clear and neatly drawn, but the authentic sketch is blurred and has indistinct and thin edges, as shown in Figure~\ref{img:brightness}. Due to the different characteristics between the training sketches and the real-world sketches, the colorization results from real-world line arts (middle images in Figure~\ref{img:brightness}) were blurred and diffused.

In order to adapt the proposed network to the real-world sketch, we trained an additional $3$ epochs with a brightness control technique~\cite{scribbler}. During the additional training stage, the brightness of the input image was scaled along $\mathcal{U}(1, 7)$, intentionally weakening the thickness and strength of the black lines so that proper segmentation was performed for an extremely faint sketch.

\section{Experiment}

\subsection{Colorization with various CVTs}

\begin{figure}[ht]
  \centering
  \includegraphics[width=0.8\columnwidth]{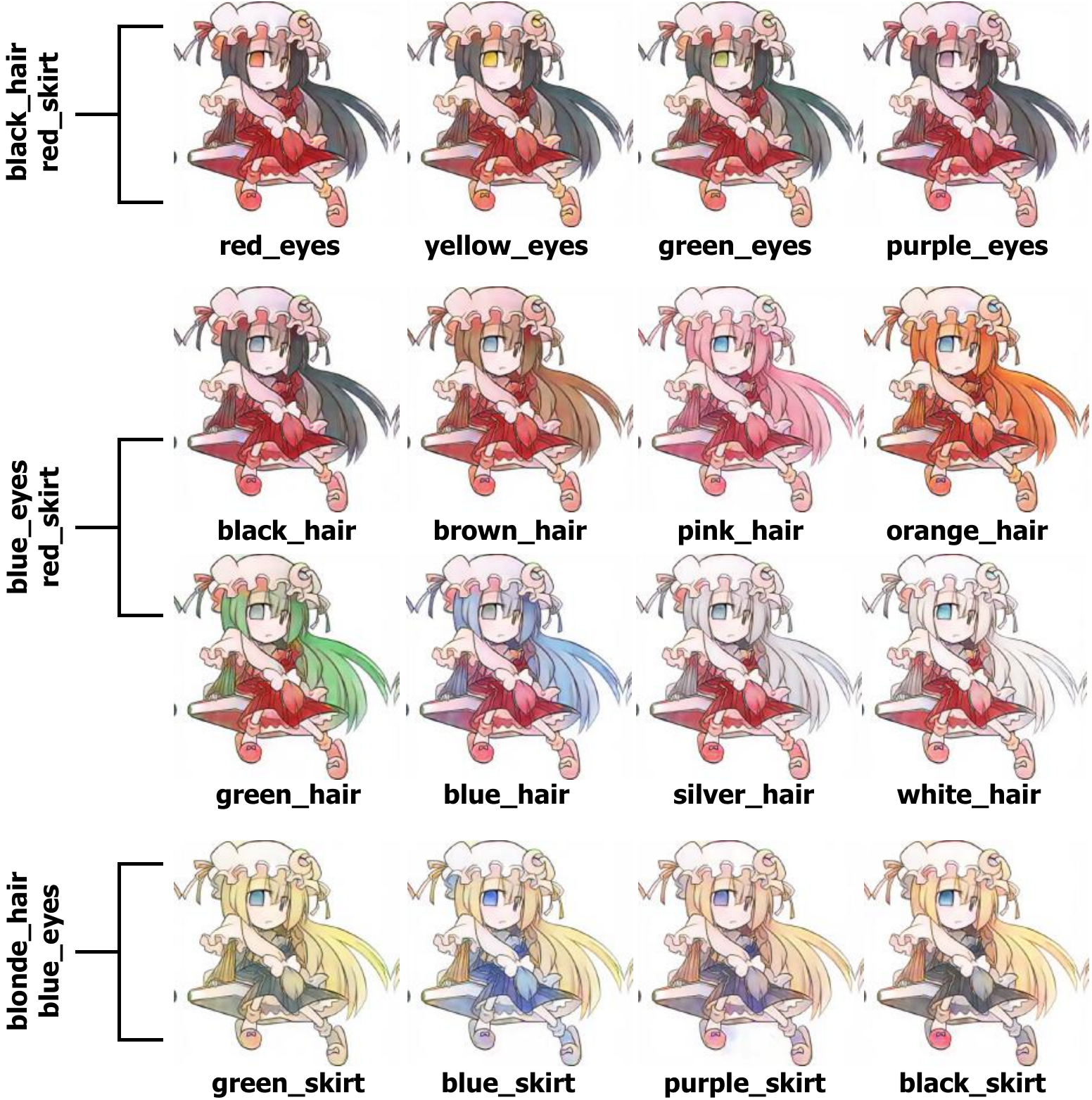}
  \caption{
Colorization results.}
  \label{img:transition}
\end{figure}

Figure~\ref{img:transition} shows the colorized images generated by Tag2Pix. Images of each row in the figure are colorized with two common CVTs and one different CVT. The images demonstrate that our network colorizes the line art naturally with various combinations of color tags.

\subsection{Comparison to other networks with user study}
\label{section_5_2}

As users will utilize automatic coloring solutions, their real judgments of the output are critical. In order to evaluate the proposed method, we conducted user studies comparing our network to other networks for sketch-based and text-based colorization. 20 people were employed offline, and they compared the quality of colorized output without any prior knowledge using a five-point Likert scale over four categories. The evaluation criteria were as follows:

\begin{itemize}
      \item \textbf{Color Segmentation} \\
      The extent to which the colors do not cross to other areas, and individual parts are painted with consistent colors.
      \item \textbf{Color Naturalness} \\
      How naturally the color conforms to the sketch. The colors should match the mood of the painting.
      \item \textbf{Color Hints Accuracy} \\
      How well the hints are reflected. Output results should have red hair if a hint for red hair is given.
      \item \textbf{Overall Quality} \\
      The overall quality of the colorized result.
\end{itemize}

\paragraph{Sketch-based network}

First, our method is compared against sketch-based line art generation methods. We chose Style2Paints (specifically style-transfer network V3)~\cite{style2paints} and PaintsChainer ~\cite{paintschainer}, which give hints using a reference image and color strokes, respectively. In Style2Paints, we created comparative images using publicly available code. In PaintsChainer, we used the service provided through the official website. We colorized 140 real-world line arts using three colorization methods. Thus, 140 test sets were made available, and each user evaluated 30 sets randomly selected out of the 140 test sets.

\begin{table}[ht]
\small
\begin{center}
\begin{tabular}{lccc}
\hline
\textbf{Categories} & \textbf{PaintsChainer (*)} & \textbf{Style2Paints} & \textbf{Tag2Pix} \\
\hline\hline
Segmentation & 2.51 & 3.51 & \textbf{3.94} \\
Naturalness & 2.44 & 3.47 & \textbf{3.91} \\
Accuracy & 3.28 & 3.73 & \textbf{3.93} \\
Quality & 2.43 & 3.47 & \textbf{3.86} \\
\hline
\end{tabular}
\end{center}
\caption{
User study of sketch-based colorization network. (*) Coloring network version was randomly selected between Tanpopo, Satsuki, and Canna.
}
\label{tab:sketchcomp}
\end{table}

\begin{figure}[ht]
  \centering
  \includegraphics[width=\columnwidth]{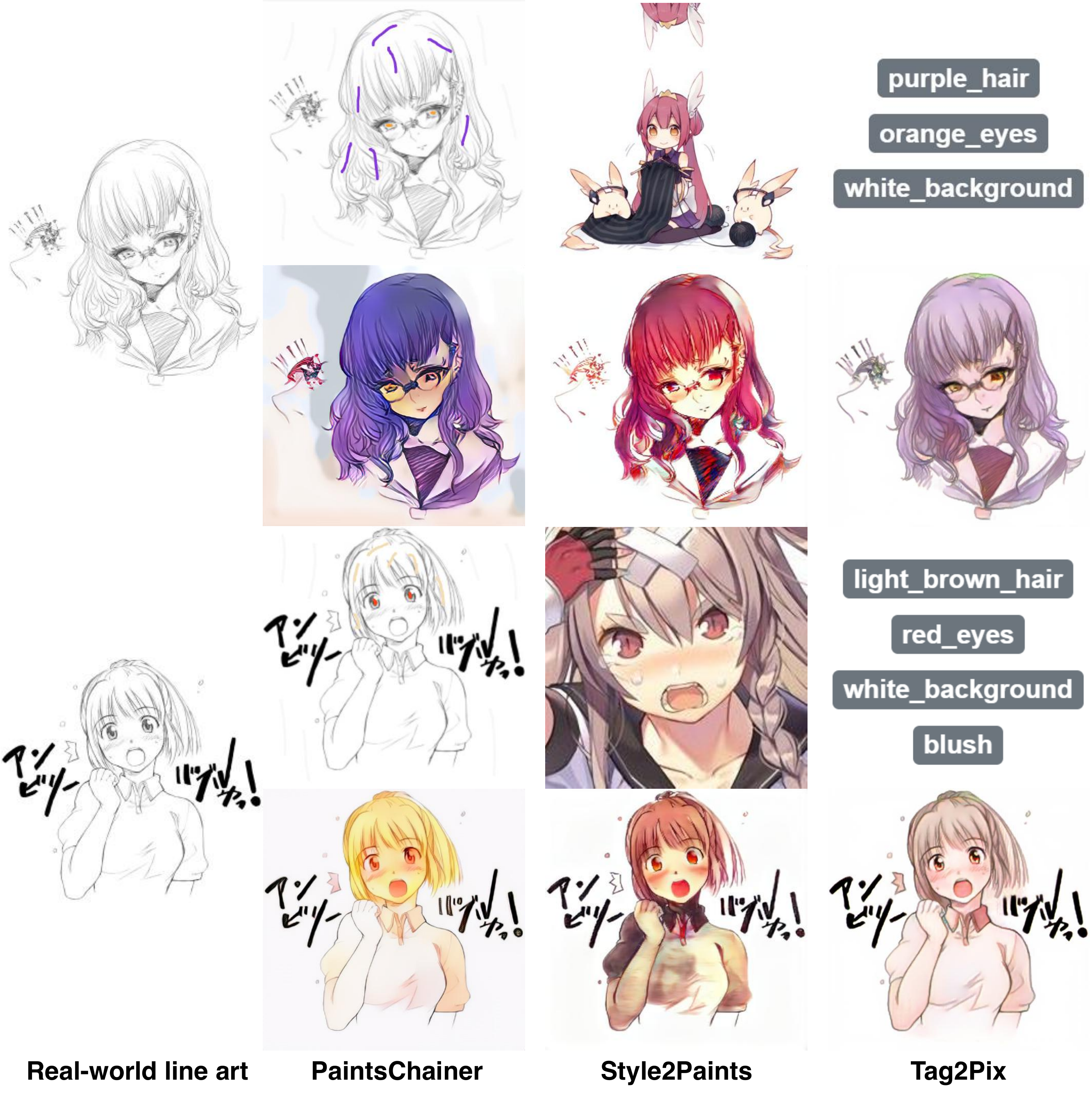}
  \caption{Comparison in sketch-based networks, PaintsChainer (left)~\cite{paintschainer} and Style2Paints (middle)~\cite{style2paints}, and ours (right). All networks painted real-world line arts. The above image of each network is a hint for colorization and the below image of each network is the colorized output. The figure shows ours has better segmentation than PaintsChainer and Style2Paints.}
  \label{img:sketch_based}
\end{figure}

Table~\ref{tab:sketchcomp} shows that Tag2Pix performed best in all evaluation metrics. In the style-transfer methods, the color is very clear and the light and shade are noticeable. However, as shown in Figure~\ref{img:sketch_based}, color mismatching and color mixing occur frequently if the reference image has a different pose relative to the line art. In PaintsChainer, the color bleeding problem is very serious. For example, eye color spreads to the face frequently due to the lack of proper segmentation for small areas. Compared to those methods, even though there is not any information about RGB values or tag location, Tag2Pix segments each part very accurately and reflects CVT hints well.

\paragraph{Text-based network}

We also conducted a comparison of Tag2Pix against the text-based colorization network. Chen \etal~\cite{chen2018language} is a conceptual design, so fair comparison is difficult because it requires a separate implementation per dataset. Additionally, SISGAN~\cite{sisgan} completely failed to colorize, despite a small input size ($74\times74$), failing to preserve the outline of the sketch, and producing a strange result. Manjunatha \etal~\cite{1804.06026} was instead selected as a comparison target, and the evaluation was conducted using the publicly available code. Because \cite{1804.06026} colorizes the image using a sentence, we converted CVTs to sentences for both training and testing, matching linguistic expression levels to ensure fair comparison. For example, \textit{red\_hair} and \textit{blue\_skirt} tags were converted to \textit{a girl with red hair wearing blue skirt}.

\begin{table}[ht]
\small
\begin{center}
\begin{tabular}{lcc}
\hline
\textbf{Categories} & \textbf{Manjunatha \etal} & \textbf{Tag2Pix} \\
\hline\hline
Segmentation & 3.16 & \textbf{4.13} \\
Naturalness & 3.46 & \textbf{4.00} \\
Accuracy & 3.27 & \textbf{3.99} \\
Quality & 3.27 & \textbf{3.86} \\
\hline
\end{tabular}
\end{center}
\caption{User study of text-based colorization network. Ours outperforms especially in segmentation.}
\label{tab:textcomp}
\end{table}

\begin{figure}[ht]
  \centering
  \includegraphics[width=\columnwidth]{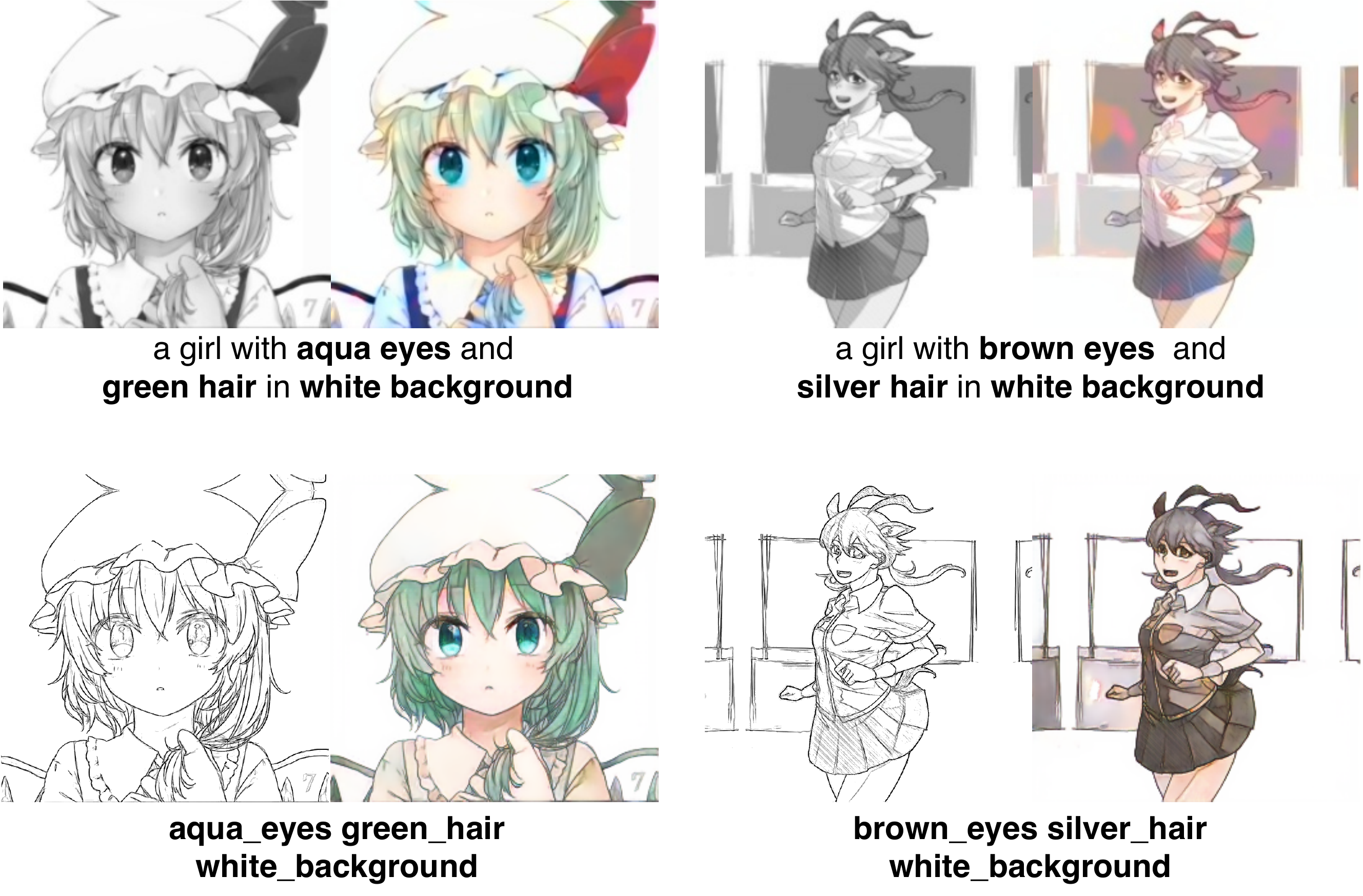}
  \caption{Comparison in text-based networks, Manjunatha \etal~\cite{1804.06026} (top), and ours (bottom). ~\cite{1804.06026} used a grayscale image and a text sentence, and ours used a line art and text tags. ~\cite{1804.06026} suffers from color bleeding and color mixing. Our colorized images have very clear lines and colors associated with the color tags.}
  \label{img:text_based}
\end{figure}

Table~\ref{tab:textcomp} shows that the proposed network significantly outperforms the baseline. As shown in Figure~\ref{img:text_based}, because~\cite{1804.06026} receives a grayscale image that preserves the shade and light of the original image, it seems plausible at a glance. However, it cannot produce a detailed segmentation of hair, eyes, etc. Our proposed network segments the specific object areas well mainly due to structural and training improvements from the CIT feature extractor, SECat, and two-step training.

Tag-based colorization has the advantage of being free from grammar and sentence structure, which significantly simplifies input encoding. Meanwhile, natural language sentences have advantages in generalization. We plan to extend the proposed approach to support natural language inputs in future work.

\subsection{Comparison between CVT embedding schemes}

Fr\'echet Inception Distance (FID)~\cite{fid} is a well-known metric for the quantitative evaluation of generated output quality. The FID quantifies the similarity between generated and ground truth images by using a pre-trained Inception-v3~\cite{inceptionv3} model to evaluate the difference of output distribution.

\begin{figure}[ht]
  \centering
  \includegraphics[width=1\columnwidth]{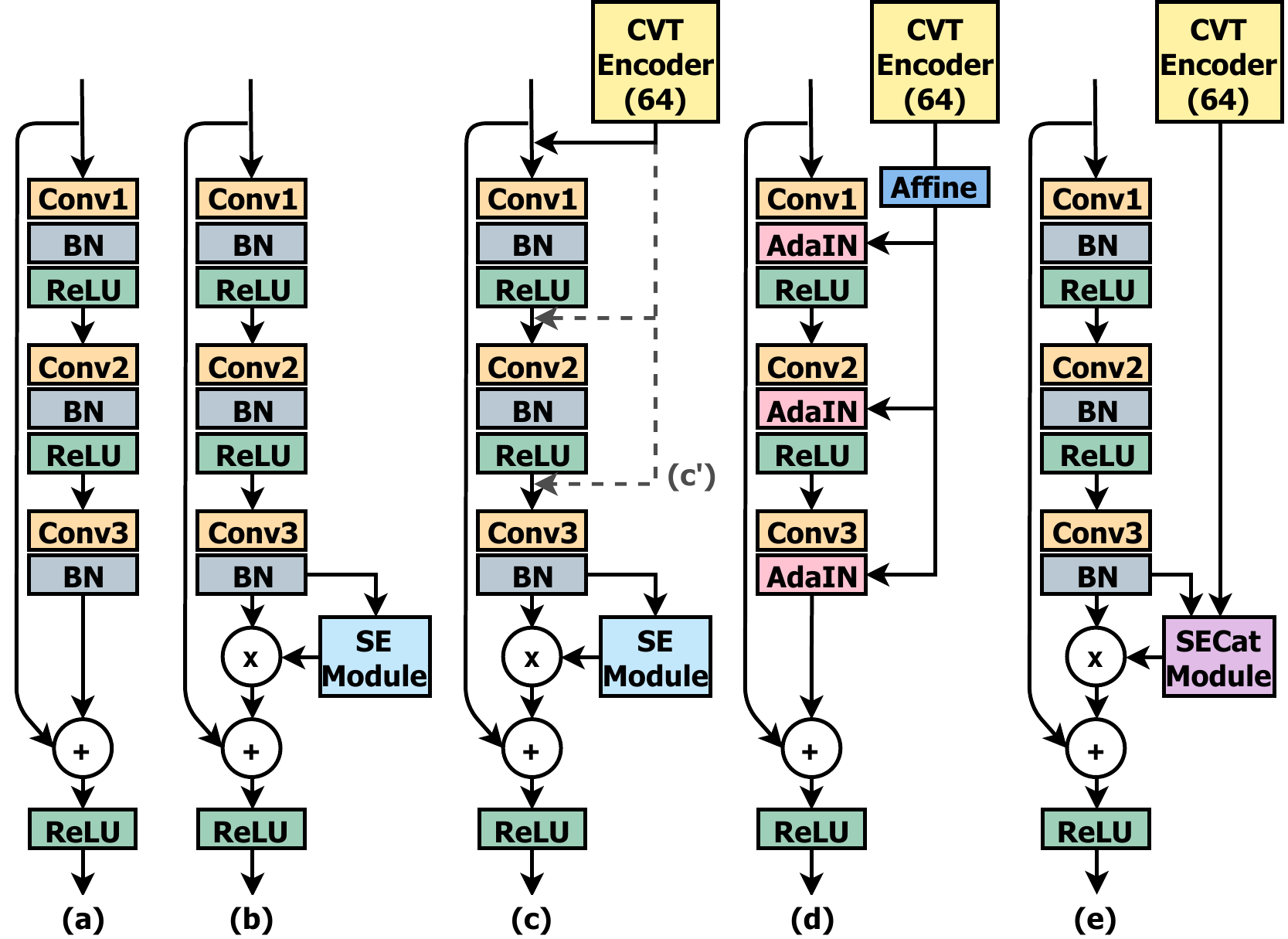}
  \caption{Diagram of each network blocks. (a) ResNeXt block, (b) SE-ResNeXt block, (c) Concatenating encoded CVT, (d) AdaIN, and (e) Ours (SECat).} 
  \label{img:SECatcomp}
\end{figure}

To show that SECat works better than other multi-label embedding methods, we embedded CVT features to our network in diverse ways and calculated the FIDs. Figure~\ref{img:SECatcomp} shows various kinds of decoding blocks in the generator. Blocks (a) and (b) are ResNeXt and SE-ResNeXt blocks, respectively. In blocks (c) and (d), CVT vector ($64$) from CVT Encoder is the same as SECat (e). The encoded vector is expanded to $H\times W\times64$ and is concatenated in front of the first conv layer of the ResNeXt block (c), or every conv layer (c$'$). (d) uses AdaIN and affine transformation as in styleGAN, and (e) is our SECat block.

We generated $6,545$ colorized images using real-world line arts and random color tags. We calculated the FID by comparing $4,127$ real illustrations in a test set. Note that the background tag is fixed as white to minimize the disturbance of background, and $10$ epochs were used for each step of the two-step training. The metric was evaluated when the network showed the lowest FID during training. The input image size was adjusted to $256\times256$ for fast comparison.

\begin{figure}[ht]
  \includegraphics[width=\columnwidth]{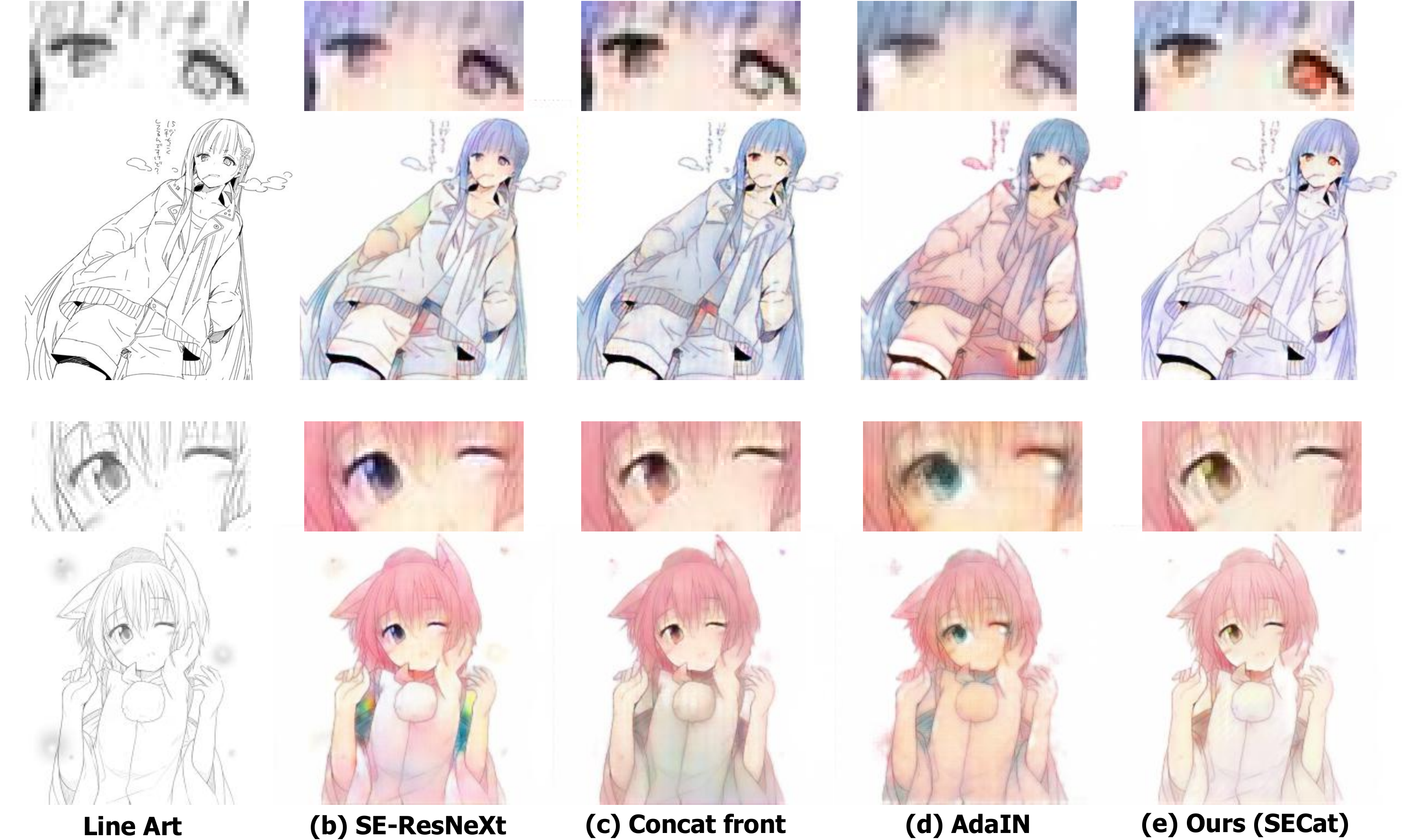}
  \caption{ Colorization output of each network block up to 20 epochs. Only our network appropriately colorizes eyes from CVT. Test inputs are real-world line arts. Top: \textit{blue\_hair, red\_eyes}, and Bottom: \textit{pink\_hair, yellow\_eyes}.} 
  \label{img:embedding_visual}
\vspace{-1mm}
\end{figure}

Table~\ref{tab:embedding} shows that our SECat block gives the lowest FID, and also has fewer parameters than other embedding methods (c) and (d). Although there is only a slight difference in FID between (c) and ours, as Figure~\ref{img:embedding_visual} shows, our method more clearly colorizes small features such as eyes.

We conducted an additional user study to compare embedding methods in more detail. All settings are almost the same as those in Section~\ref{section_5_2}, but we hired 27 new users and adopted $6,545$ images used for calculating FIDs. As shown in Table~\ref{tab:embedding_userstudy}, (b) merges CVT features for spatial features once with the deepest decoder block, and is inferior to (c) and (e), which merge additional light CVT features in every decoder block. By incorporating CVT information into the network and utilizing that information to emphasize the features, (e) SECat dominated all evaluation criteria.

\begin{table}[ht]
\small
\begin{center}
\begin{tabular}{lcc}
\hline
\textbf{Block} & \textbf{FID} & \textbf{Params} \\
\hline\hline
(a) ResNeXt & 52.49 & 13.85M \\
(b) SE-ResNeXt & 45.48 & 15.71M \\
(c) Concat front & 39.69 & 16.03M \\
(c$'$) Concat all & 47.08 & 17.37M \\
(d) AdaIN & 66.39 & 16.51M \\
(e) SECat (ours) & \textbf{39.21} & 15.81M \\
\hline
\end{tabular}
\end{center}
\caption{FIDs and parameter counts of each generator network. Ours gives the best result in FID and the smallest amount of additional parameters.}
\label{tab:embedding}
\vspace{-2mm}
\end{table}

\begin{table}[ht]
\small
\begin{center}
\begin{tabular}{lccc}
\hline
\textbf{Categories} & \textbf{(b)} & \textbf{(c)} & \textbf{(e) SECat} \\
\hline\hline
Segmentation & 2.85 & 3.29 & \textbf{3.51} \\
Naturalness & 2.85 & 3.40 & \textbf{3.60} \\
Accuracy & 2.75 & 3.51 & \textbf{3.60} \\
Quality & 2.78 & 3.30 & \textbf{3.54} \\
\hline
\end{tabular}
\end{center}
\caption{User study of CVT embedding schemes. (b), (c) and (e) correspond to SE-ResNeXt, Concat front and SECat in Table~\ref{tab:embedding}, respectively.}
\label{tab:embedding_userstudy}

\vspace{-4mm}
\end{table}

\subsection{Ablation studies and colorization for sequential sketches}

We conducted extensive studies analyzing the effect of network components and loss terms, and performed colorization using consistent tags for sequential sketches on video. Check the supplementary material for details.

\section{Conclusion}

In this paper, we presented a novel GAN-based line art colorization called Tag2Pix which produces high-quality colored images for given line art and color tags, and trained the network with an introduced Tag2Pix dataset. The proposed SECat architecture properly colorizes even small features, and the proposed two-step training scheme shows that pre-training the network for segmentation is a prerequisite to learning better colorization. Various user studies and experiments show that Tag2Pix outperforms the existing methods in terms of segmentation, naturalness, representation of color hints, and overall quality.

\vspace{-1.6mm}
\section*{Acknowledgement}
This work was supported by National Research Foundation of Korea (NRF-2016M3A7B4909604) and Samsung Electronics.

{\small
\bibliographystyle{ieee_fullname}

}

\newpage
\newpage

\clearpage


\section*{Supplementary material}

# Supplementary material for "Tag2Pix: Line Art Colorization Using Text Tag With SECat and Changing Loss"

Hyunsu Kim*, Ho Young Jhoo*, Eunhyeok Park, and Sungjoo Yoo

Seoul National University

{gustnxodjs, mersshs, eunhyeok.park, sungjoo.yoo}@gmail.com

## 1. Ablation study with component and loss

We conducted three types of ablation studies to demonstrate the effects of each component and loss proposed in our Tag2Pix network. The experimental setting is the same as that in Section 5.3 of the paper. We calculated FIDs [3] while ablating each component and loss in our network. Tables 1, 2, and 3 show the contribution of each component and loss.

| Network No. | N1 | N2 | N3 | N4 | N5 |
|---|---|---|---|---|---|
| CIT feature | - | O | O | O | O |
| Guide decoder | - | - | O | O | O |
| SECat | - | - | - | O | O |
| Two-step | - | - | - | - | O |
| FID | 57.81 | 52.78 | 48.74 | 42.29 | **39.21** |

Table 1. Incremental component ablation study. The rightmost column is our best network with all components.

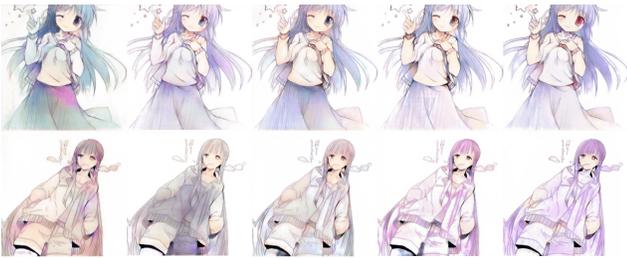

Figure 1. Images correspond to N1 to N5 from Table 1, left to right. Row 1 images were colorized with *blue_hair* and *red_eyes*; row 2 images were colorized with *purple_hair* and *pink_eyes*.

Table 1 shows that the FID decreases as more components are added. That is, colorization results gradually follow the color distribution of original images. As shown in Figure 1, from left to right, the quality of the image increases and the color of each segment becomes clearer and

---
*equal contribution

more similar to the given CVTs. Without CIT and the guide decoder (N1, N2), segmentation misses and color bleeding can occur. Without SECat and two-step training, as in the case of N3, the colors corresponding to the CVTs are not rendered well. N4, which is not trained by two-step training, is still insufficient, although some similar colors are painted in a particular position. After all components are used, the desired colors are appropriately painted in the desired areas.

| Network No. | N6 | N7 | N8 | N4 | N5 |
|---|---|---|---|---|---|
| CIT feature | - | O | O | O | O |
| Guide decoder | O | - | O | O | O |
| SECat | O | O | - | O | O |
| Two-step | O | O | O | - | O |
| FID | 44.20 | 46.41 | 45.48 | 42.29 | **39.21** |

Table 2. Component ablation study removing one component at a time.

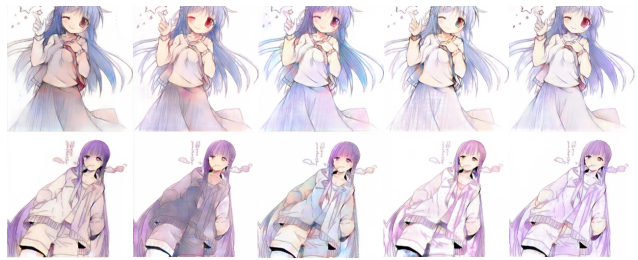

Figure 2. Ablation study colorization removing one component at a time. Images correspond to N6 to N5 from Table 2, left to right. Row 1 images were colorized with *blue_hair* and *red_eyes*; row 2 images were colorized with *purple_hair* and *pink_eyes*.

To show each component's effectiveness in more detail, we conducted additional experiments removing one component at a time to present quantitative and qualitative results. Table 2 and Figure 2 show that each component individually contributed to improving quality. The CIT feature ex-

tractor provides color invariant information to the generator as a hint. Thus, the generator can localize well (N5 vs. N6 in Table 2 and the fifth vs. first column in Figure 2). The network needs all of the CIT features, SECat, and two-step training to colorize small feature like eyes (N5, N7). Without SECat, the colors are not uniform (N8). So far, we have seen shortages in coloring when not using any of the components.

| Network No. | N4 | N5 | N5 | N5 |
|---|---|---|---|---|
| Adversarial loss | O | - | O | O |
| Classification loss | - | O | O | O |
| Reconstruction loss | O | O | - | O |
| FID | 54.19 | 54.33 | -(*) | **39.21** |

Table 3. Loss ablation study. The rightmost column is the loss combination we used. (*) Failed to generate human-like image.

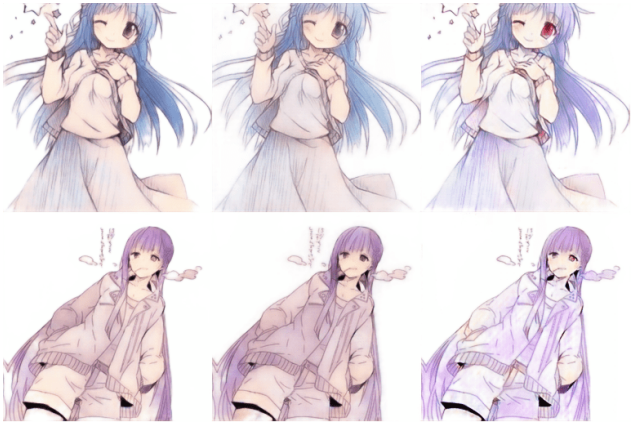

Figure 3. Colorization results of loss ablation study. Without classification loss (left), without adversarial loss (middle) and all loss contained (right). Row 1 images were colorized with *blue_hair* and *red_eyes*; row 2 images were colorized with *purple_hair* and *pink_eyes*.

In Table 3 and Figure 3, we used the same metric as in the component ablation studies. Note that, without classification loss, it is impossible to use a two-step method, so the ablation study with classification loss used the N4 network. Without classification loss or adversarial loss, the tone of color becomes dull overall and the small features, such as eyes, do not have the desired color. Eyes are instead generally painted in similar colors overall and are unnatural and lacking shade. Among loss functions, reconstruction loss (L1 loss) is the most important because it makes the network learn the overall distribution of real images so that the network can colorize somewhat naturally without classification loss or adversarial loss. However, our network colorizes small features such as eyes with the color specified by CVTs only when both adversarial loss and classification loss are in place.

## 2. User study details

20 users were recruited offline as evaluators without prior knowledge of our work. They evaluated networks in two sections, between sketch-based networks and text-based networks. In each section, they were given an hour to evaluate 30 sets of test images. Four categories of evaluation metrics were rated with a five-point Likert scale as follows:

- **Color Segmentation**
  The extent to which the colors do not cross to other areas, and individual parts are painted with consistent colors.

- **Color Naturalness**
  How naturally the color conforms to the sketch. The colors should match the mood of the painting.

- **Color Hints Accuracy**
  How well the hints are reflected. Output results should have red hair if a hint for red hair is given.

- **Overall Quality**
  The overall quality of the colorized result.

### 2.1. Sketch-based networks

We chose PaintsChainer [10], which is famous for colorizing an image with color strokes, and Style2Paints [7, 11], which is state-of-the-art in line art colorization. In PaintsChainer, we used the service provided through the official website. In Style2Paints, we created comparative images using the publicly available code (only V3 is publicly available). We prepared 140 real-world line arts and made 140 test sets. Each user evaluated 30 sets randomly selected out of the 140 test sets. Each test set consisted of PaintsChainer [10], Style2Paints [7], and Tag2Pix (ours).

| Network | PaintsChainer | Style2Paints | Tag2Pix |
|---|---|---|---|
| Acc. Mean | 3.28 | 3.73 | **3.93** |
| Nat. Mean | 2.44 | 3.47 | **3.91** |
| Quality Mean | 2.43 | 3.47 | **3.86** |
| Seg. Mean | 2.51 | 3.51 | **3.94** |
| Acc. Std. | 1.29 | 1.06 | **0.98** |
| Nat. Std. | 1.23 | 1.24 | **1.05** |
| Quality Std. | 1.19 | 1.15 | **0.98** |
| Seg. Std. | 1.28 | 1.24 | **1.03** |

Table 4. Mean and standard deviation of score in sketch-based networks.

Table 4 shows that Tag2Pix has the highest average scores and the lowest standard deviation in all evaluation metrics, indicating that our colorization network generates the most stable and highest quality images from line arts. Figure 4 also shows a box chart of the comparison results.

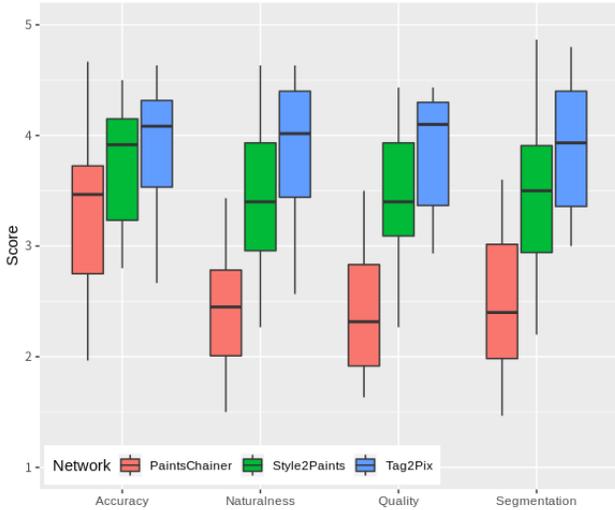

Figure 4. The box chart of sketch-based networks.

As PaintsChainer generally had much lower scores than Tag2Pix, we focused on comparing Style2Paints and Tag2Pix. Figure 8 shows evaluation images for which Tag2pix received a much higher rating than Style2Paints. Style2Paints images have very high contrast such that they look exaggerated. Additionally, strange colors spread on the arms and legs due to poor segmentation. Tag2pix is generally good at segmentation and provided natural colorization. However, Figure 9 shows evaluation images for which Style2Paints received a much higher rating than Tag2pix. As Style2Paints images have better distinct color and shade, some images that had proper segmentation got better scores. If the refinement stage of Style2Paints is added to our method, the image quality could be improved.

## 2.2. Text-based networks

There are comparable networks which use text as a hint for colorization. Chen *et al*. [1] colorizes a grayscale image using a text sentence hint. SISGAN [2] changes the color of certain parts of a color image as described in the sentence. However, Chen *et al*. [1] is a conceptual design, so fair comparison is difficult because it requires a separate implementation per dataset as done in the paper. In addition, SISGAN [2] completely failed to colorize, despite a small $74 \times 74$ RGB input because it was not suitable for our dataset and task. It failed to preserve the outline of the sketch and produced a strange result, thus we exclude them as a comparison target.

Instead, Manjunatha *et al*. [8] was selected as a comparison target, and the evaluation was conducted using the publicly available code [9]. Because Manjunatha *et al*. [8] colorizes the image using a sentence, we converted CVTs to sentences with fixed structure for both training and testing.

| Network | Manjunatha *et al.* | Tag2Pix |
|---|---|---|
| Acc. Mean | 3.27 | **3.99** |
| Nat. Mean | 3.46 | **4.00** |
| Quality Mean | 3.27 | **3.86** |
| Seg. Mean | 3.16 | **4.13** |
| Acc. Std. | 1.09 | **0.95** |
| Nat. Std. | 1.17 | **0.99** |
| Quality Std. | 1.09 | **0.93** |
| Seg. Std. | 1.17 | **0.93** |

Table 5. Mean and standard deviation of score in text-based networks.

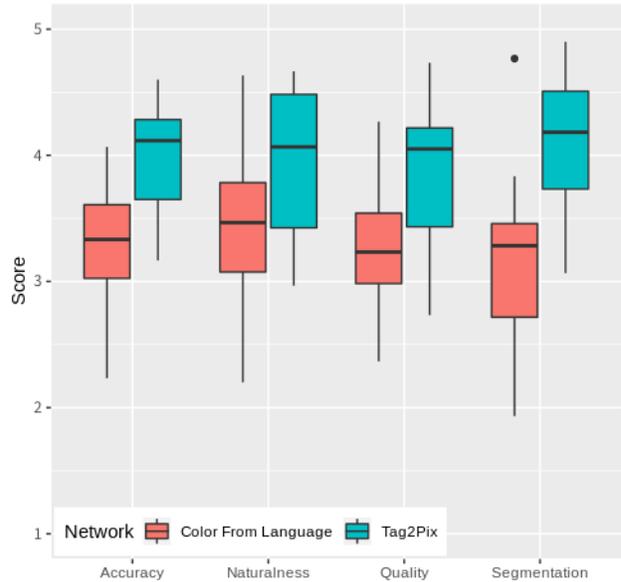

Figure 5. The box chart of text-based networks.

For example, *red_hair* and *blue_skirt* tags were converted to *a girl with red hair wearing blue skirt*.

As shown in Table 5 and Figure 5, even though our network's input image (line art) has much less information than Manjunatha *et al*. [8]'s input image (grayscale), ours has much better average scores and lower standard deviation in all evaluation metrics.

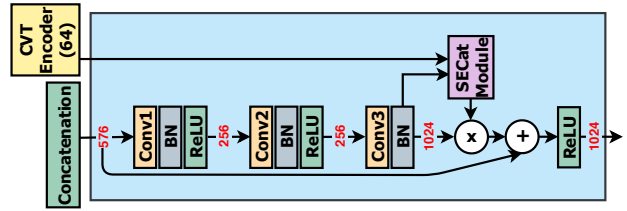

Figure 6. Proposed SECat-ResNeXt block structure.

## 3. SECat

Figure 6 illustrates the SECat-ResNeXt block structure. It shows how the decoder module, combining the Concatenation block, SECat-ResNeXt block, and PixelShuffle, decodes input features to output features with different dimensions using three convolution layers, *e.g.*, $576 \rightarrow 1024$ dimensions in the generator's first decoder module (Figure 3 in the paper).

As shown in Figure 5 in the paper, SECat module input is the output from the CVT encoder, and utilized to re-weight the residual block output feature maps. Figure 6 shows that CVT encoder output is only used by the SECat module, rather than by the residual block convolution layers. The SECat exploits weight re-balancing following SENet [4] and styleGAN [6], incorporating hints from the CVT encoder to re-calibrate the ResNeXt block output feature maps.

Compared to previous concepts [5, 12] for incorporating global information through concatenation, SECat can improve coloring quality by utilizing CVT information when re-calibrating output feature maps. Thus, SECat both incorporates CVT information into the network (see Figure 3 in the paper) and utilizes it to emphasize the features (shown in the Figure 6 above and Figure 5 in the paper).

In particular, SECat helps to significantly improve generation quality by enhancing fine detail colorization, including small objects such as eyes, compared with the concatenation method in Figure 11(c), as mentioned in Section 5.3 of the paper.

However, the FIDs in Table 3 of the paper struggle to indicate this improvement because the metric mainly focuses on overall colorization distribution. To show the SECat module's effectiveness on fine details, we conducted an additional user study to compare generated image output quality between embedding methods. We also released the SECat code[1] so that readers can reproduce the results.

### 3.1. Additional user study in CVT embedding schemes.

The experimental environment is similar to that in Section 2, employing 27 people without prior knowledge, and comparing three networks, each of which has a different hint embedding method. The evaluation time, method and number of evaluation sets are the same as in Section 2, but the evaluation data was the $6,545$ images used in Section 5.3 of the paper. We compared the three networks with the lowest FID in Table 3 of the paper, those being *SE-ResNeXt*, *Concat front*, and *SECat* (ours).

As shown in Table 6 and Figure 7, *SECat* is superior in all criteria. The mean is high, and the standard deviation is low. Comparing the results with *SE-ResNeXt* and other

---

[1]https://github.com/blandocs/Tag2Pix

networks, the method of consistently inserting information about hints into all decoders is overwhelmingly effective. There is little FID difference between *Concat front* and *SECat* in Table 3 of the paper, but there is a somewhat meaningful difference in the user study. We have shown through user study that SECat is superior to other embedding methods.

| Network | SE-ResNeXt | Concat front | SECat (ours) |
|---|---|---|---|
| Acc. Mean | 2.75 | 3.51 | **3.60** |
| Nat. Mean | 2.85 | 3.40 | **3.60** |
| Quality Mean | 2.78 | 3.30 | **3.54** |
| Seg. Mean | 2.85 | 3.29 | **3.51** |
| Acc. Std. | 1.06 | 1.04 | **1.00** |
| Nat. Std. | 1.18 | 1.06 | **1.00** |
| Quality Std. | 1.17 | 1.08 | **1.01** |
| Seg. Std. | 1.19 | 1.13 | **1.08** |

Table 6. Mean and standard deviation of score of different CVT embedding schemes.

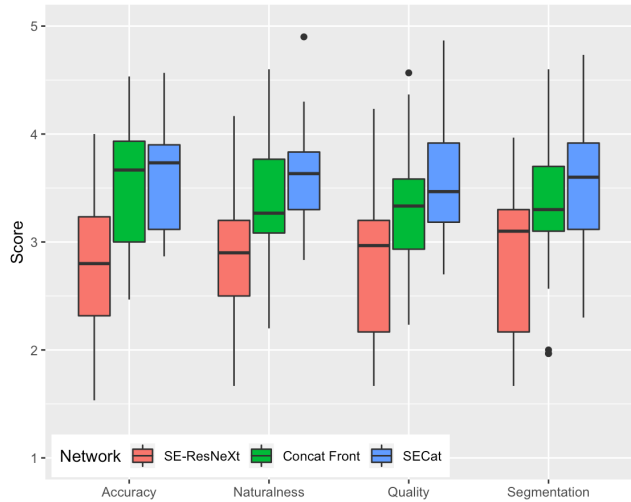

Figure 7. The box chart of different CVT embedding schemes.

## 4. More results and future work

Our proposed network has the potential for colorizing any pose in consecutive sketches with a single tag set, which makes Tag2Pix suitable for animation colorization. In Figures 10 and 11, we can see the feasibility of this idea. In the style-transfer colorization, which colorizes a line art with a reference image, Style2Paints cannot colorize properly unless the human pose of the line art is similar to the reference image. The color of the head and skin is especially misplaced even though we used the same character as a reference image in Figure 10. Through our method, we can colorize each part of the character properly regardless of the pose. This can be used as a tool in character design or

in the pre-production stage of the 2D animation production process.

We generated additional examples that can be seen in Figures 12 and 13. We changed the color of shirt, skirt, legwear, skin, and blush, as well as eyes and hair that were previously compared in the paper.

## Acknowledgement

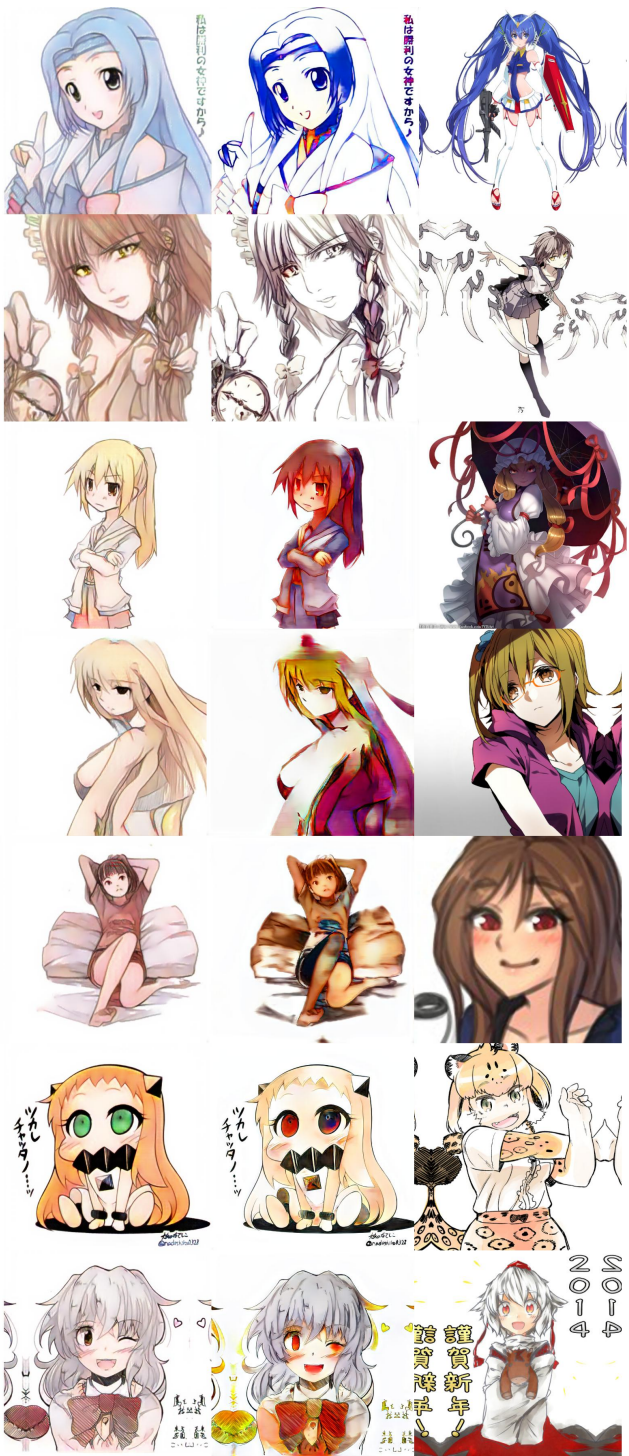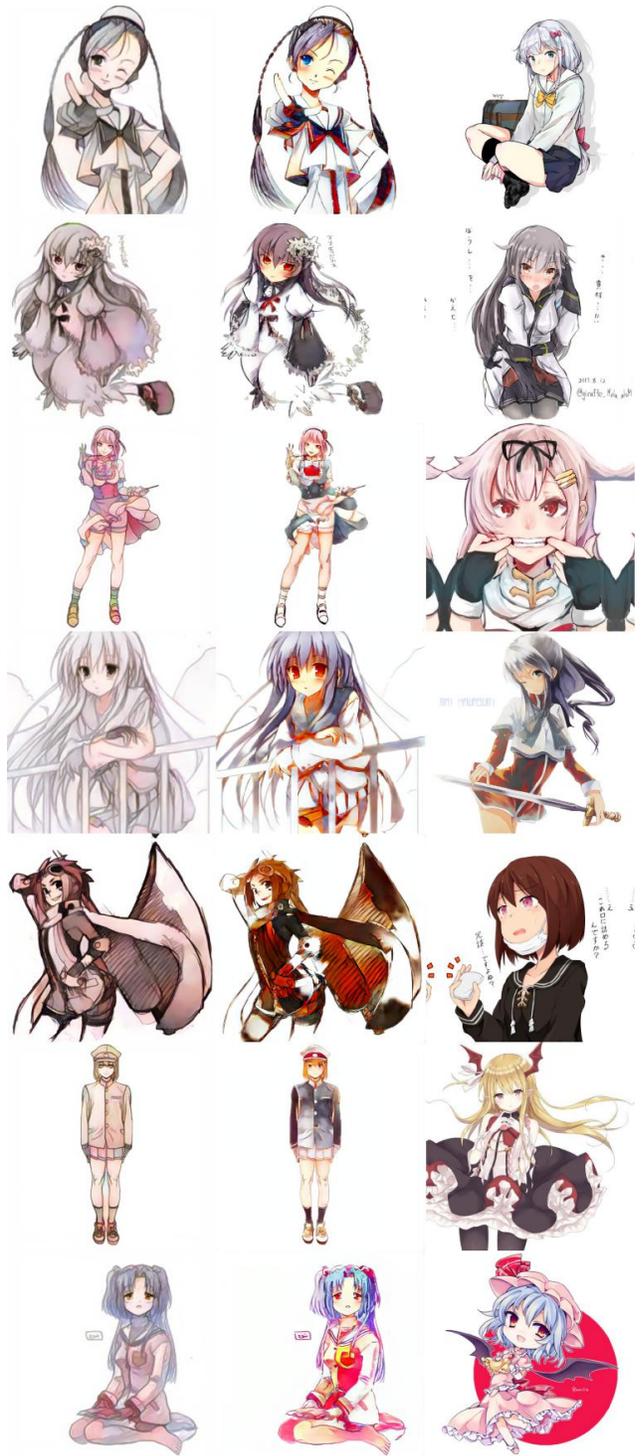

Figure 8. The images that Tag2pix (Ours) received a much higher rating than Style2Paints. Tag2pix colorization result (left), Style2Paints colorization result (middle) and Style2Paints reference images (right).

Figure 9. The images that Style2Paints received a much higher rating than Tag2pix (Ours). Tag2pix colorization result (left), Style2Paints colorization result (middle) and Style2Paints reference images (right).

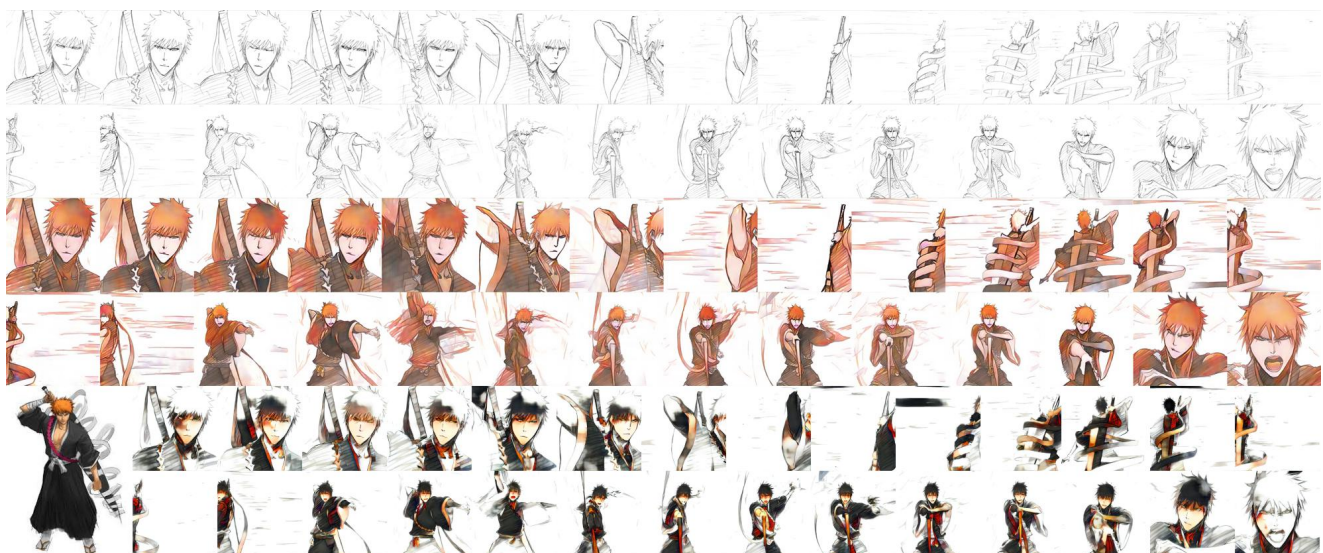

Figure 10. Animation colorization sample. The first and second rows are line arts in animation. The third and fourth rows are Tag2Pix (Ours) colorization results. We only give three tags to each frame, *white_background, orange_hair* and *black_dress*. The fifth and sixth rows are Style2Paints [7] colorization results. We used the leftmost image for reference.

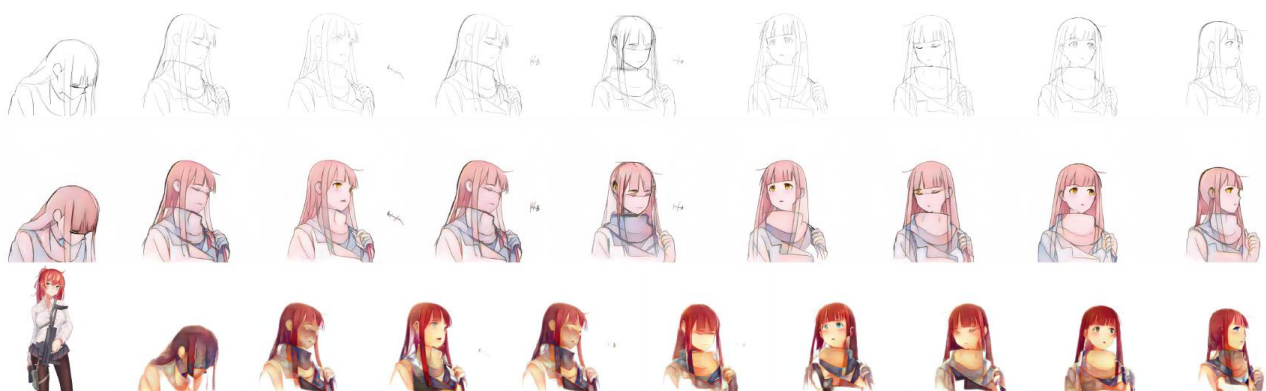

Figure 11. Animation colorization sample. The first row is line arts in animation. The second row is Tag2Pix (Ours) colorization results. We only give four tags to each frame, *white_background, pink_hair, yellow_eyes*, and *white_shirt*. The third row is Style2Paints [7] colorization results. We used the leftmost image for reference.

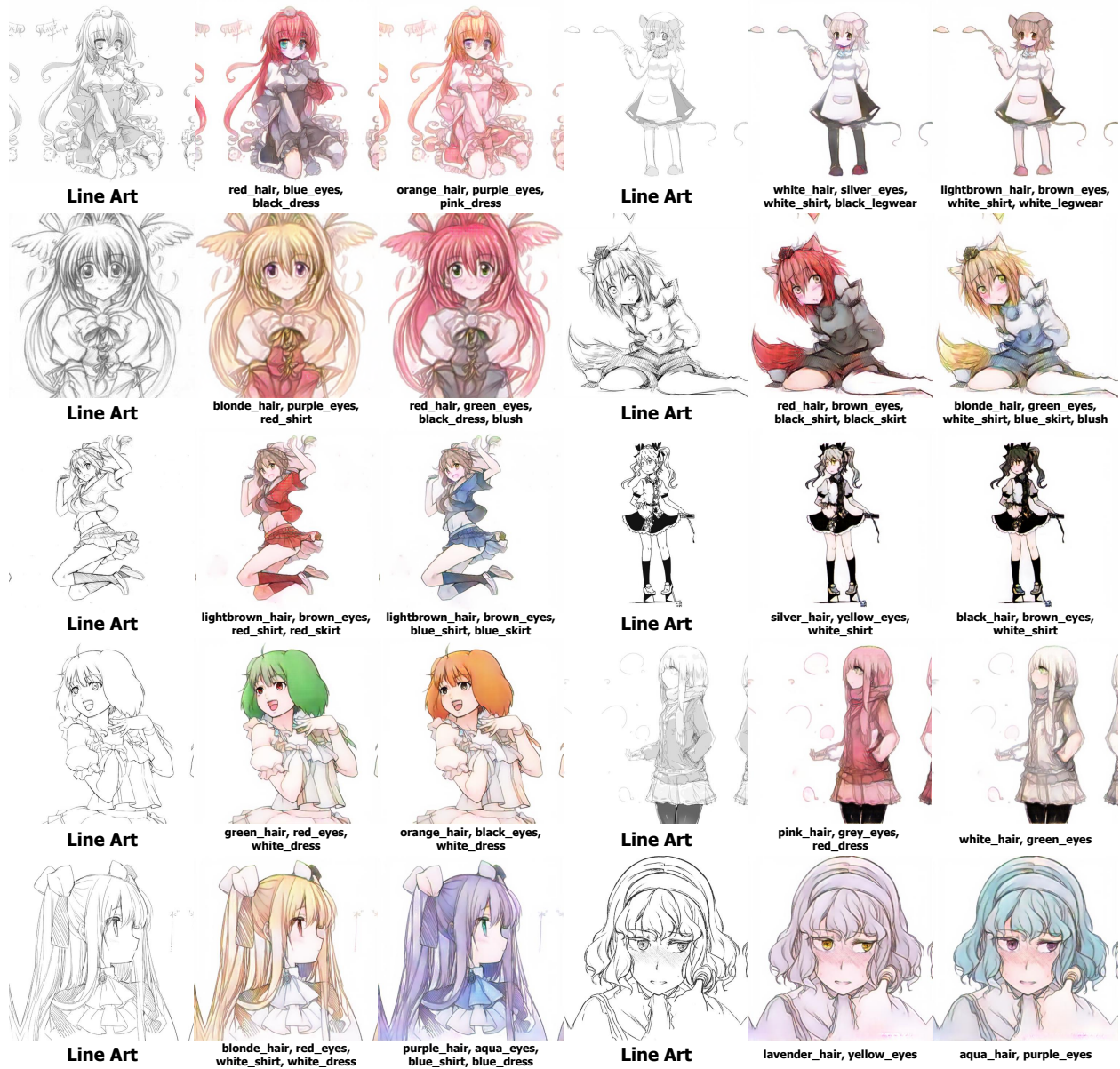

Figure 12. Results of our network. All images are colorized with the tag *white_background*.

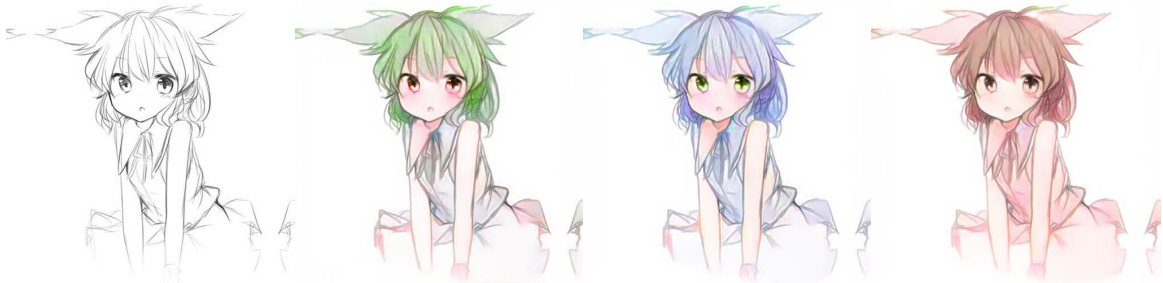
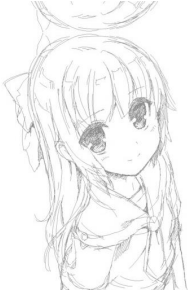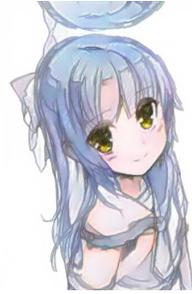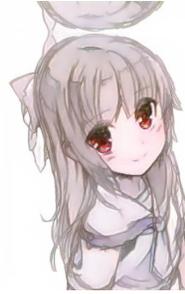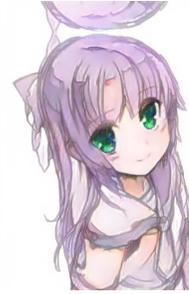
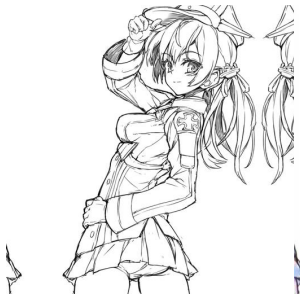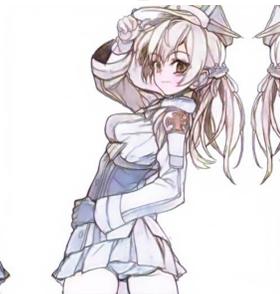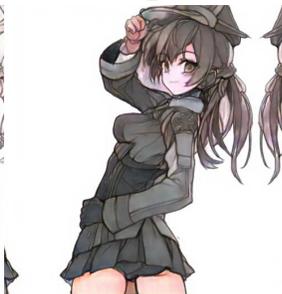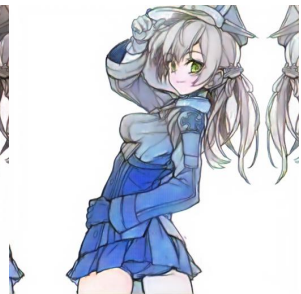
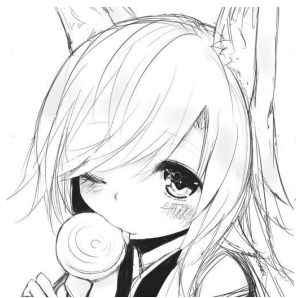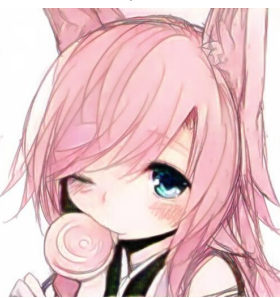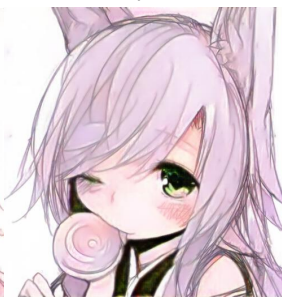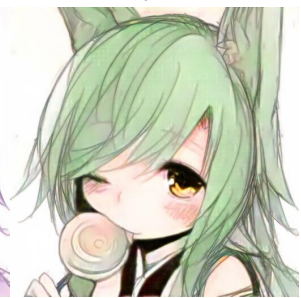
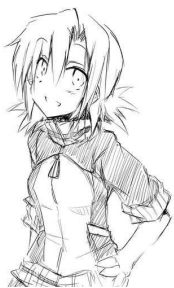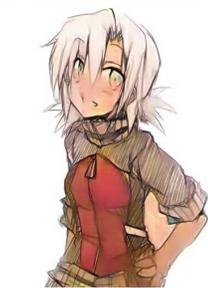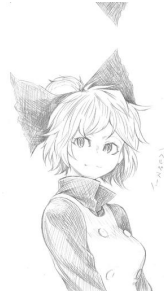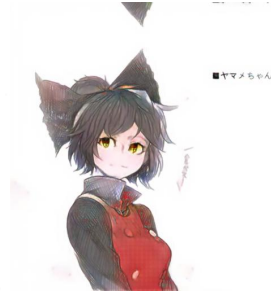

Figure 13. Results of our network. All images are colorized with the tag *white_background*.



\end{document}